\documentclass[10pt,twocolumn,letterpaper]{article}

\usepackage{iccv}
\makeatletter
\@namedef{ver@everyshi.sty}{}
\makeatother
\usepackage{times}
\usepackage{epsfig}
\usepackage{graphicx}
\usepackage{amsmath}
\usepackage{amssymb}

\usepackage{booktabs}
\usepackage{tabularx}
\usepackage{xcolor}
\newcommand{\new}[1]{\textcolor{orange}{{\bf}{\em #1}{\bf}}}
\newcommand{\newseb}[1]{\textcolor{green}{{\bf}{\em #1}{\bf}}}
\usepackage{subcaption}
\usepackage{algorithm}
\usepackage{algpseudocode}

\usepackage{amssymb}
\usepackage{amsthm}
\newtheorem{theorem}{Theorem}[section]
\usepackage{mathtools}
\usepackage[thinc]{esdiff}
\usepackage{todonotes}
\usepackage{comment}
\usepackage[symbol]{footmisc}

\usepackage[pagebackref=true,breaklinks=true,letterpaper=true,colorlinks,bookmarks=false]{hyperref}

\usepackage{cleveref}

 \iccvfinalcopy 

\def\iccvPaperID{12506} 

\ificcvfinal\fi

\begin{document}

\title{SAMSON: Sharpness-Aware Minimization Scaled by Outlier Normalization for Improving DNN Generalization and Robustness}

\author{Gonçalo Mordido\footnotemark\;\textsuperscript{\rm 1,2}, Sébastien Henwood\footnotemark[1]\;\textsuperscript{\rm 2}, Sarath Chandar\textsuperscript{\rm 1,2,3}, François Leduc-Primeau\textsuperscript{\rm 2}\\
\textsuperscript{\rm 1}Mila - Quebec AI Institute, \textsuperscript{\rm 2}Polytechnique Montreal, \textsuperscript{\rm 3}Canada CIFAR AI Chair\\
}

\maketitle
\footnotetext[1]{Equal contribution.}
\ificcvfinal\thispagestyle{empty}\fi

\begin{abstract}
   Energy-efficient deep neural network (DNN) accelerators are prone to non-idealities that degrade DNN performance at inference time. To mitigate such degradation, existing methods typically add perturbations to the DNN weights during training to simulate inference on noisy hardware. However, this often requires knowledge about the target hardware and leads to a trade-off between DNN performance and robustness, decreasing the former to increase the latter. In this work, we show that applying sharpness-aware training, by optimizing for both the loss value and loss sharpness, significantly improves robustness to noisy hardware at inference time without relying on any assumptions about the target hardware. In particular, we propose a new adaptive sharpness-aware method that conditions the worst-case perturbation of a given weight not only on its magnitude but also on the range of the weight distribution. This is achieved by performing sharpness-aware minimization scaled by outlier minimization (SAMSON). Our approach outperforms existing sharpness-aware training methods both in terms of model generalization performance in noiseless regimes and robustness in noisy settings, as measured on several architectures and datasets.
\end{abstract}


\section{Introduction}
\label{sec:introduction}

The success of deep neural networks (DNNs) has also been accompanied by an increase in training complexity and computational demands, prompting efficient DNN designs~\cite{mordido2019instant,mordido2020monte,mordido21_interspeech}. However, with the slowing down of Moore's law and the ending of Dennard scaling, power consumption is now the key design constraint for DNN accelerators, which calls for new hardware and algorithms. Emerging approaches, such as in-memory computing, are promising directions to improve the energy consumption and throughput of existing DNNs~\cite{sze2020efficient}. This is particularly important in computer vision applications that have low-energy or high-throughput requirements.

Despite their appeal, highly efficient hardware implementations are often prone to variabilities~\cite{10.1145/2463209.2488867} which perturb the DNN weights and lead to a degradation of DNN performance~\cite{joshi2020accurate,kern2022memse,10.3389/fncom.2021.675741,tambe2020edgebert}. The main approach for improving the robustness of DNNs has been to apply weight perturbations during training~\cite{henwood2020layerwise,8792205,8993573,joshi2020accurate}. However, such approaches typically rely on noise simulations from the target hardware to which the DNN will be deployed. 
Moreover, existing robustness methods provide a trade-off between DNN performance and DNN robustness, decreasing the former to increase the latter.

The goal of this work is to increase both model robustness and performance without relying on any noise simulations from the target hardware. By doing so, we do not compromise the applicability of our approach, neither by reducing the original DNN performance nor by tailoring it to a specific hardware design. To achieve this, we propose to apply sharpness-aware minimization methods during training to promote accurate DNN inference after deployment on noisy, yet energy-efficient, hardware.

The benefit of converging to a smoother loss landscape has been primarily tied to improving generalization performance~\cite{hochreiter1994simplifying,keskar2016large,DR17,neyshabur2017exploring,chaudhari2017entropysgd,DBLP:conf/uai/IzmailovPGVW18}. With this goal in mind, Foret \etal \cite{foret2021sharpnessaware} recently proposed sharpness-aware minimization (SAM) by minimizing both the loss value and loss sharpness within a maximization region around each parameter during training. By showing a high correlation between loss sharpness and test performance, SAM has ignited several follow-up works since its proposal. Particularly, adaptive SAM (ASAM)~\cite{kwon2021asam} reformulated sharpness to be invariant to weight scaling by conditioning the neighborhood region of each weight based on its magnitude.

In this work, we propose to perform sharpness-aware minimization scaled by outlier normalization (SAMSON) to increase not only the performance but also the robustness of DNNs. SAMSON reformulates adaptive sharpness to consider not only the weight magnitude but also the range of the weight distribution. By promoting adaptivity based on the outlier weights, we achieve better generalization performance on multiple DNN architectures (ResNet-18, ResNet-32, ResNet-50, VGG-13, MobileNetV2, and DenseNet-40), datasets (CIFAR-10, CIFAR-100, and ImageNet), and learning regimes (training from scratch and finetuning). 

Additionally, we show that SAMSON's sharpness measure has a high correlation with model robustness. In other words, SAMSON's objective may be used during training in combination with existing robustness techniques to increase DNN robustness at inference time. This is observed on a generic noise model as well as on accurate noise simulations from real hardware. The presented improvements in terms of test accuracy in both noiseless and noisy regimes showcase the extensive practicality of our approach.


\section{Related work}

The deployment of pre-trained models on noisy hardware for highly efficient inference is known to introduce non-idealities. This is caused by noise inherent to the device~\cite{programming_read_noise} such as programming noise after weight transfer to the target hardware and read noise every time the programmed weights are accessed. Without robustness measures, such hardware noise significantly hinders the performance of neural networks. To promote robustness after deployment in noisy hardware at inference time, existing methods typically inject noise or faults to DNN weights during training \cite{ambrogio2018equivalent,10.3389/fncom.2021.675741,li2019long,conductance_drift,mackin2019weight}. In particular, adding weight noise \cite{joshi2020accurate} and promoting redundancy by performing aggressive weight clipping \cite{stutz2021bit} have been shown to be effective methods for increasing DNN robustness. However, existing robustness methods often lead to a decrease of DNN performance for promoting robustness. Moreover, they typically rely on noise measurements from the target hardware to improve the performance and robustness trade-off.

Sharpness-aware training has recently gathered increased interest~\cite{Sun_Zhang_Ren_Luo_Li_2021,Jiang*2020Fantastic,foret2021sharpnessaware,chen2022when}. Particularly, SAM has sparked a lot of new follow-up works due to the significant increase in generalization performance presented in the original paper. Variants mainly focus on increasing the efficiency \cite{du2022sharpness,du2021efficient,zhou2021delta,liu2022towards,zhao2022ss}, performance \cite{zhuang2022surrogate,kim2022fisher,kwon2021asam}, or understanding \cite{pmlr-v162-andriushchenko22a} of sharpness-aware training. Efforts have also been made to extend SAM to specific use-cases such as quantization-aware training \cite{liu2021sharpness} or data imbalance settings \cite{liu2021selfsupervised}. Several works~\cite{kwon2021asam,dinh2017sharp} have also highlighted the importance of scale-invariant sharpness measures, including in the context of model robustness against adversarial examples~\cite{Stutz_2021_ICCV}.

In a similar vein to our work, Sun \etal \cite{Sun_Zhang_Ren_Luo_Li_2021} recently related the sharpness of the loss landscape with robustness to adversarial noise perturbations. This was further observed by Kim \etal \cite{kim2022fisher}. We follow this under-explored research direction and provide an in-depth study on the effect of loss sharpness in robustness against noisy hardware. Stutz \etal \cite{Stutz_2021_ICCV} also recently studied the flatness of the (robust) loss landscape on the basis of adversarial training with perturbed examples~\cite{madry2018towards}. In particular, they tackle the problem of robust overfitting~\cite{206180}, \textit{i.e.} having high robustness to adversarial examples seen during training but generalizing poorly to new adversarial examples at test time, through the lens of flat minima. Here, we focus on the problem of improving robustness against noisy weights (rather than adversarial examples) at inference time.

\section{Sharpness-aware minimization (SAM)}

The goal of sharpness-aware minimization or SAM is to promote a smoother loss landscape by optimizing for both the loss value and loss sharpness during training. Generally speaking, given a parameter $w$, the goal is to find a region in the loss landscape where not only does $w$ have a low training loss $L$ but also do its neighbor points. Considering the $L_2$ norm and discarding the regularization term in the original algorithm for simplicity, SAM uses the following objective:
\begin{equation}
    L_\text{SAM}(w) = \min_{w} \max_{\|\epsilon\|_2 \leq \rho} L(w + \epsilon),
    \label{eq:sam}
\end{equation}
where the size of the neighborhood region is defined by a sphere with radius $\rho$ and the optimal $\hat{\epsilon}$ may be efficiently estimated via a first-order approximation, leading to:
\begin{equation}
    \pmb{\epsilon}^*_\text{SAM}(\pmb{w}) = \rho \: \frac{\nabla L(\pmb{w})}{||\nabla L(\pmb{w})||_2} \,.
\end{equation}
By building on the strong correlation between sharpness and generalization performance, SAM is generally used in practice to achieve better test performance. However, there are two main drawbacks. The first is that, despite its efficiency in estimating the worst-case weight perturbations, SAM's objective requires two forward passes for every backward pass. To mitigate this additional time complexity, the authors propose to leverage distributed training. Another drawback of SAM is that the sharpness calculation is not independent from weight scaling. This allows the manipulation of sharpness values by applying scaling operators to the weights such that weight values change without altering the model's final prediction \cite{dinh2017sharp,Stutz_2021_ICCV}.

\subsection{Adaptive sharpness-aware minimization (ASAM)}

To tackle the scale variance issue, adaptive sharpness-aware minimization or ASAM was proposed by Kwon \etal\cite{kwon2021asam}. By taking into account scaling operators that do not change the model's loss, ASAM creates a new notion of adaptive sharpness that is invariant to parameter scaling, contrarily to SAM. This is reflected in ASAM's objective:
\begin{equation}
    L_\text{ASAM}(w) = \min_{w} \max_{\|\epsilon / |w| \|_2 \leq \rho} L(w + \epsilon),
    \label{eq:asam}
\end{equation}
where $|w|$ represents the absolute value of a given weight $w$. With ASAM, different neighborhood sizes are applied to different weights, depending on their magnitude; high-magnitude weights withstand higher perturbations than low-magnitude weights. This adaptive sharpness formulation also leads to a change in the neighborhood shape, which is now ellipsoidal instead of spherical. The worst-case perturbation $\pmb{\epsilon}^*_\text{ASAM}$ is defined as
\begin{equation}
    \pmb{\epsilon}^*_\text{ASAM}(\pmb{w}) = \rho \: \frac{w^2\nabla L(w)}{||w \nabla L(w)||_2} \,. \quad \text{(elementwise ops.)}
    \label{eq:eps_asam}
\end{equation}
In practice, the adaptive sharpness that ASAM introduced shows a higher correlation with generalization performance and overall improved convergence by using larger maximization regions for larger weights. 

\section{Sharpness-aware minimization scaled by outlier normalization (SAMSON)}

In this work, we propose a novel sharpness- and range-aware method called sharpness-aware minimization scaled by outlier normalization or SAMSON. In essence, our approach considers not only the weight magnitude but also the range of the weight distribution to determine the perturbation $\epsilon$ of a weight $w$. Conditioning sharpness by weight magnitude and the dynamic range of the weight distribution leads to the neighborhood sizes being normalized across all layers. This is particularly important when training with batch normalization, since the scales of the weight distributions across different layers may greatly differ leading to a discrepancy in the applied weight perturbations across the entire network. 

We propose to take into account the outlier weight, \textit{i.e.} the maximum absolute weight of a given layer, by simply scaling the effective neighborhood size of a weight $w$ by the $p$-norm of all the weights $\pmb{w}$:
\begin{equation}
    L_\text{SAMSON}(w) = \min_{w} \max_{\|\epsilon \|\pmb{w}\|_p / |w| \|_2 \leq \rho} L(w + \epsilon),
    \label{eq:samson}
\end{equation}
which leads to the following per-weight worst-case perturbation:
\begin{equation}
    \pmb{\epsilon}^*_\text{SAMSON}(\pmb{w}) = \rho \: \frac{(w\|\pmb{w}\|_p^{-1})^2\nabla L(w)}{||w\|\pmb{w}\|_p^{-1} \nabla L(w)||_2} \, . \quad \text{(elementwise ops.)}
    \label{eq:eps_samson}
\end{equation}
We note that the $p$-norm affects the impact of outlier weights in the applied worst-case perturbation. In our study, we experiment with using $p = \{2, \infty\}$. For ease of presentation, we often refer to SAMSON variants with $p=2$ and $p=\infty$ as SAMSON$_2$ and SAMSON$_\infty$, respectively, throughout the paper.

\begin{figure}[t]
     \centering
     \includegraphics[width=0.48\textwidth]{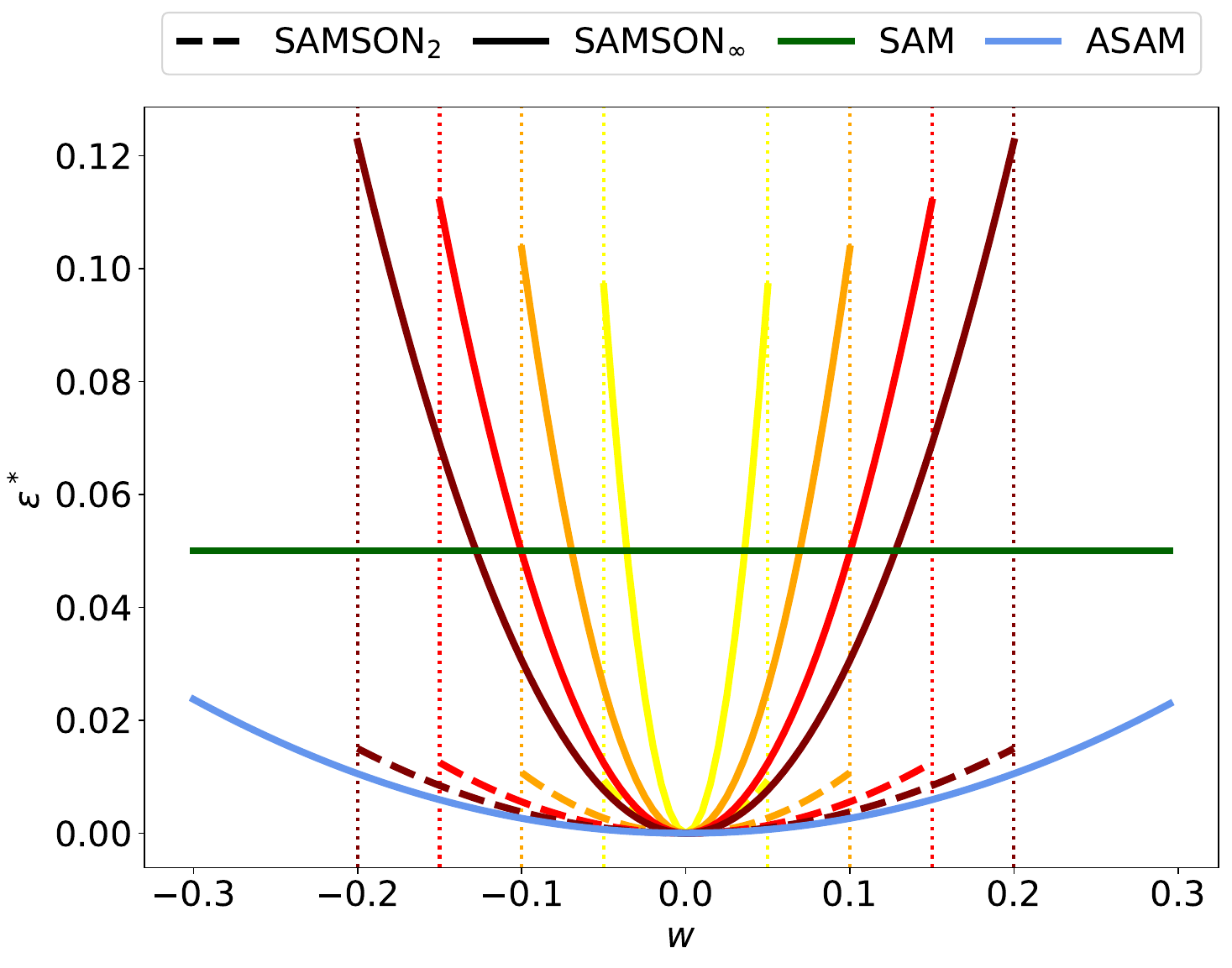}
     \caption{Worst-case perturbations of SAMSON, ASAM, and SAM.}
     \label{fig:samson}
\end{figure}

We illustrate the applied worst-case perturbation considering a given weight value with SAMSON, ASAM, and ASAM in \cref{fig:samson}, assuming $\nabla L(w)=1$ for simplicity. To showcase that our method is adaptive not only to the weight magnitude but also to the weight range, we apply different symmetric ranges to the original weight distribution. Particularly, we restrict the weights to be within [-$c$, $c$], with $c \in \{0.1, 0.15, 0.2\}$. We see that $\epsilon^*_\text{SAM}$ is independent of both the weight value and weight range and as a result, is represented as a straight line which is defined solely by $\rho$. Since $\epsilon^*_\text{ASAM}$ depends on $\rho$ and the weight magnitude, larger weights are more perturbed. However, since ASAM is independent of the weight range, there is no change in ASAM's perturbations when changing the range of the weight distribution. On the other hand, SAMSON is both range- and weight magnitude-dependent, taking into account the weight value, $\rho$, and outlier weights for its perturbations. This results in the observed changes in $\epsilon^*_\text{SAMSON}$ over the different ranges, with SAMSON$_2$ putting less emphasis on outlier weights and SAMSON$_\infty$ emphasizing them.

Despite not depending on any form of weight clipping, our approach is inherently suited to be used in combination with methods that restrict the weight distribution range. For example, training with aggressive weight clipping~\cite{stutz2021bit} to improve robustness at inference time. When applying weight clipping, $c$ is defined as the clipping range. In the case of aggressive weight clipping, the weights are forced to be inside a small range to promote robustness: \textit{i.e.} $c \in \mathbb{R}: 0 < c < 1$. A pseudo-code implementation of SAMSON combined with aggressive weight clipping is presented in \cref{alg:samson}. We would like to highlight that performing weight clipping is an optional step in our algorithm, and should only be performed if one is interested in improving DNN robustness in noisy regimes and not DNN performance in the noiseless setting.


\begin{algorithm}
\caption{SAMSON combined with weight clipping.}
\begin{algorithmic}
\Require initial weight $\pmb{w_0}$, aggressive clipping range $c$, learning rate $\alpha$, neighborhood size $\rho$, norm $p$
\State $\pmb{w} \gets \pmb{w_0}$
\While{not converged}
\State Sample minibatch $s$
\State $\pmb{\epsilon} = \rho \: \frac{(w\|\pmb{w}\|_p^{-1})^2\nabla L_s(w)}{||w\|\pmb{w}\|_p^{-1} \nabla L_s(w)||_2}$ \Comment{elementwise ops.}
\State $\pmb{w} \gets \pmb{w} - \alpha \nabla L_s(\pmb{w} + \pmb{\epsilon}) $ \Comment{weight update}
\State $\pmb{w} \gets \text{clip}(\pmb{w},c)$ \Comment{weight clipping (optional)}
\EndWhile
\end{algorithmic}
\label{alg:samson}
\end{algorithm}


\begin{table*}
\begin{center}
\begin{tabular}{|l|c|c|c|c|c|}
\hline
Method & ResNet-34 & ResNet-50 & MobileNetV2 & VGG-13 & DenseNet-40\\
\hline\hline
SGD & $95.84_{\pm 0.13}$ & $94.36_{\pm 0.09}$ & $94.62_{\pm 0.06}$ & $94.19_{\pm 0.04}$ & $91.76_{\pm 0.11}$\\
SAM & $95.80_{\pm 0.07}$ & $94.24_{\pm 0.13}$ & $94.91_{\pm 0.07}$ & $94.52_{\pm 0.07}$ & $92.27_{\pm 0.30}$\\
ASAM & $95.85_{\pm 0.22}$ & $94.42_{\pm 0.57}$ & $95.37_{\pm 0.04}$ & $94.68_{\pm 0.07}$ & $\pmb{92.57_{\pm 0.06}}$\\
\hline
SAMSON$_2$ & $\pmb{95.96_{\pm 0.34}}$ & $\pmb{95.09_{\pm 0.21}}$ & $95.29_{\pm 0.17}$ & $\pmb{94.73_{\pm 0.12}}$ & $92.54_{\pm 0.14}$\\
SAMSON$_\infty$ & $95.76_{\pm 0.29}$ & $\pmb{94.94_{\pm 0.09}}$ & $\pmb{95.41_{\pm 0.09}}$ & $94.66_{\pm 0.02}$ & $92.49_{\pm 0.13}$\\
\hline
\end{tabular}
\end{center}
\caption{Generalization performance (test accuracy \%) of the different methods on several models trained on CIFAR-10. We present the mean and standard deviation over 3 runs.}
\label{tab:cifar_10_generalization}
\end{table*}

\begin{table*}
\begin{center}
\begin{tabular}{|l|c|c|c|c|c|}
\hline
Method & ResNet-34 & ResNet-50 & MobileNetV2 & VGG-13 & DenseNet-40\\
\hline\hline
SGD & $74.32_{\pm 1.32}$ & $74.35_{\pm 1.23}$ & $75.44_{\pm 0.07}$ & $72.78_{\pm 0.22}$ & $68.52_{\pm 0.25}$\\
SAM & $75.62_{\pm 0.33}$ & $75.36_{\pm 0.01}$ & $76.81_{\pm 0.18}$ & $73.86_{\pm 0.40}$ & $69.14_{\pm 0.36}$\\
ASAM & $76.91_{\pm 0.44}$ & $77.88_{\pm 0.85}$ & $77.28_{\pm 0.10}$ & $74.12_{\pm 0.01}$ & $70.21_{\pm 0.25}$\\
\hline
SAMSON$_2$ & $\pmb{77.68_{\pm 0.57}}$ & $\pmb{78.22_{\pm 0.67}}$ & $77.24_{\pm 0.13}$ & $\pmb{74.77_{\pm 0.23}}$ & $69.94_{\pm 0.36}$\\
SAMSON$_\infty$ & $\pmb{77.60_{\pm 0.78}}$ & $77.81_{\pm 1.32}$ & $\pmb{77.61_{\pm 0.23}}$ & $\pmb{74.59_{\pm 0.15}}$ & $\pmb{70.34_{\pm 0.37}}$\\
\hline
\end{tabular}
\end{center}
\caption{Generalization performance (test accuracy \%) of the different methods on several models trained on CIFAR-100. We present the mean and standard deviation over 3 runs.}
\label{tab:cifar_100_generalization}
\end{table*}

\section{Generalization performance}
\label{sec:generalization_performance}

We will first focus on DNN performance in the noiseless setting, which is the most common concern in practice. We analyze the generalization performance of SAMSON, ASAM, SAM, and SGD on ResNet-34~\cite{he2016deep}, ResNet-50, MobileNetV2~\cite{sandler2018mobilenetv2}, VGG-13~\cite{simonyan2014very}, and DenseNet-40~\cite{huang2017densely} models trained on CIFAR-10 and CIFAR-100~\cite{krizhevsky2009learning}. We trained each model for 200 epochs with a batch size of 128, starting with a learning rate of 0.1 and dividing it by 10 every 50 epochs. 
We tested the different sharpness-aware training variants using the default neighborhood sizes proposed in the original papers of SAM and ASAM. In particular, we set $\rho = 0.05$ and $\rho = 0.5$ for SAM and ASAM on CIFAR-10, respectively, and $\rho = 0.1$ and $\rho = 1.0$ for SAM and ASAM on CIFAR-100, respectively. To promote a direct method comparison, we report the same default $\rho$ as ASAM suggested for our SAMSON variants.

The test accuracy comparisons between the different methods on CIFAR-10 and CIFAR-100 are shown in tables \ref{tab:cifar_10_generalization} and \ref{tab:cifar_100_generalization}, respectively. 
We see that at least one of our variants (SAMSON$_2$ or SAMSON$_\infty$) consistently outperforms the SGD, SAM, and ASAM variants in terms of generalization performance. The only instance where this is not observed is when using DenseNet-40 trained on CIFAR-10, with ASAM outperforming SAMSON's best variant by a negligible margin ($+0.03\%$). Nevertheless, we note that SAMSON$_\infty$ outperforms ASAM using VGG-13 on CIFAR-100 by $+0.13\%$. However, we would like to note that VGG-13 is the worst-performing DNN architecture across both datasets which lessens the significance of these results. Overall, both SAMSON variants with $p=2$ and $p=\infty$ achieve the best test accuracies, with the best performing $p$ being dataset and architecture dependent.

To further expand our exploration of different models and datasets, we finetuned a ResNet-18 model pre-trained on ImageNet~\cite{russakovsky2015imagenet} provided by PyTorch. More specifically, we first initialized our models with the weights of the aforementioned ResNet-18 model and then finetuned them for a total of 10 epochs with a batch size of 400, a learning rate of 0.001, and a weight decay of 0.0001. Since no default $\rho$ is reported in the original ASAM's paper for finetuning on ImageNet, we iterate over different neighborhood ranges (details are provided in the appendix) and report the best performing $\rho$ for SAM, ASAM, and SAMSON. In the end, the best performances were obtained using $\rho = 0.05$ for SAM, $\rho = 0.2$ for SAMSON, and $\rho = 0.5$ for ASAM. Generalization results are presented in \cref{tab:imagenet_generalization}. We observe an improvement in terms of top-1 and top-5 test accuracy when using SAMSON with both the $p=2$ and $p=\infty$ compared to the other methods.

Generally, we observe that the adaptive sharpness formulation of ASAM to weight magnitude usually outperforms SAM's sharpness in regard to generalization performance. However, this is not always observed, particularly on ImageNet, with SAM slightly outperforming ASAM's test accuracy. By also adapting the neighborhood region to depend on the weight distribution range, SAMSON's variants improve generalization performance across the tested models and datasets.
These results highlight the generality of our approach, which may not only be used to achieve more robust DNNs (as will be discussed in the next sections) but also to increase generalization performance in standard training settings. Moreover, the ImageNet results also suggest that SAMSON is a suitable method to perform finetuning on large models, which further expands its benefits in practice.

\begin{table}
\begin{center}
\begin{tabular}{|l|c|c|c|c|}
\hline
Method & \multicolumn{2}{c|}{ResNet-18}\\
 & top-1 & top-5 \\
\hline\hline
SGD & $69.758$ & $89.078$\\
SAM & $70.356$ & $89.480$\\
ASAM & $70.348$ & $89.428$\\
\hline
SAMSON$_2$ & $\pmb{70.358}$ & $\pmb{89.486}$\\
SAMSON$_\infty$ & $\pmb{70.366}$ & $\pmb{89.504}$\\
\hline
\end{tabular}
\end{center}
\caption{Generalization performance (top-1 and top-5 test accuracy \%) of the different methods with ResNet-18 finetuned on ImageNet. SGD represents the model performance before finetuning.}
\label{tab:imagenet_generalization}
\end{table}

\begin{figure*}[ht]
     \centering
     \begin{subfigure}[b]{0.33\textwidth}
         \centering
         \includegraphics[width=\textwidth]{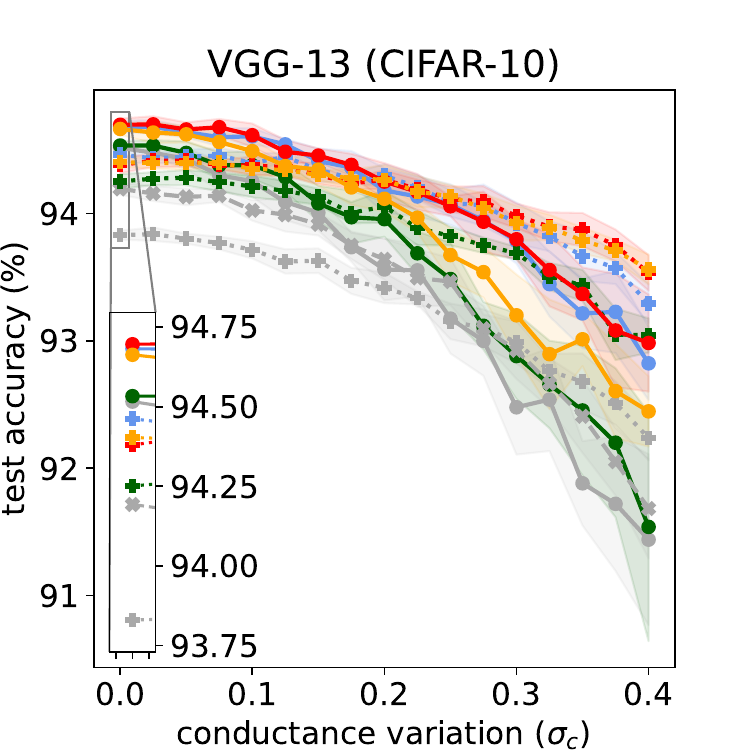}
     \end{subfigure}
     \hfill
     \begin{subfigure}[b]{0.33\textwidth}
         \centering
         \includegraphics[width=\textwidth]{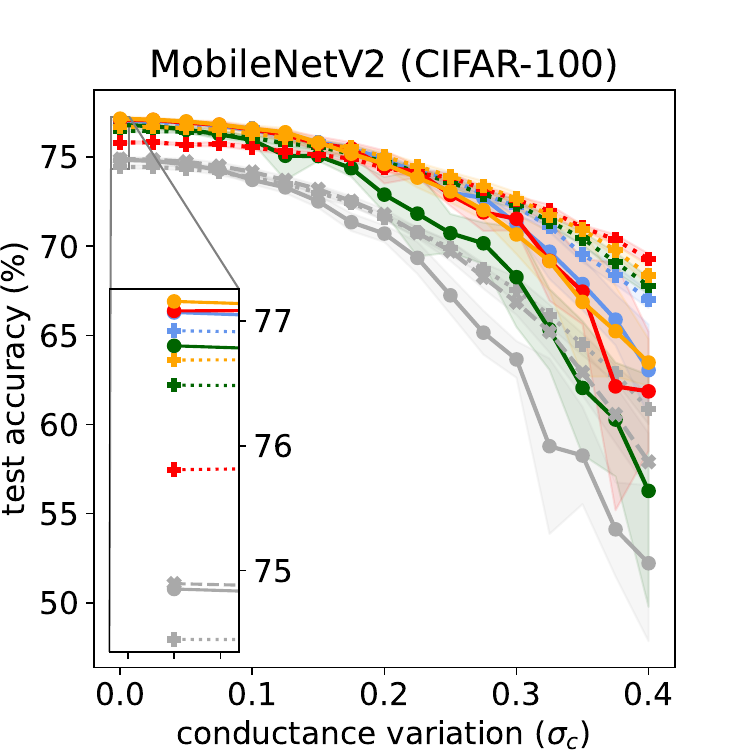}
     \end{subfigure}
     \hfill
     \begin{subfigure}[b]{0.33\textwidth}
         \centering
         \includegraphics[width=\textwidth]{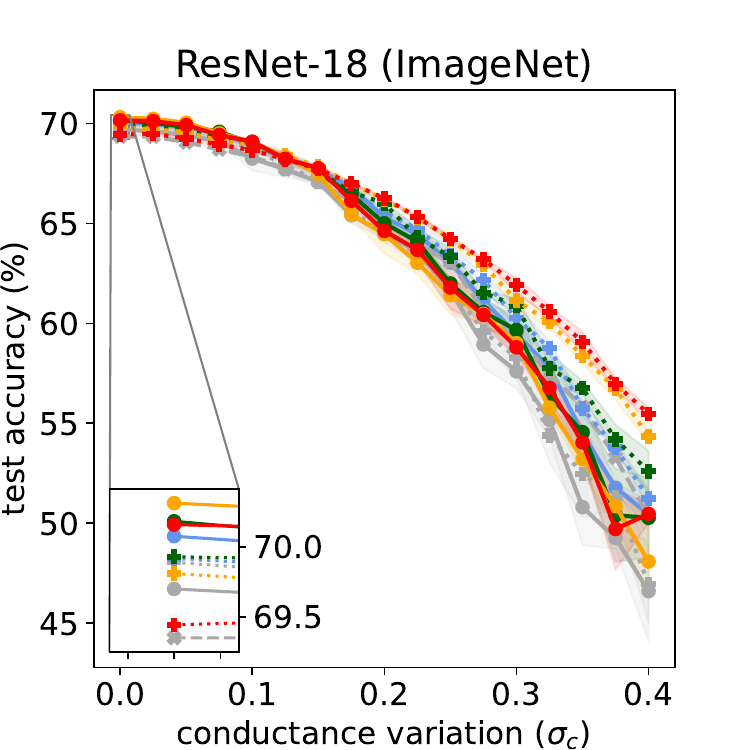}
     \end{subfigure}
     \hfill     
     \begin{subfigure}[b]{0.99\textwidth}
        \centering
        \includegraphics[width=\textwidth]{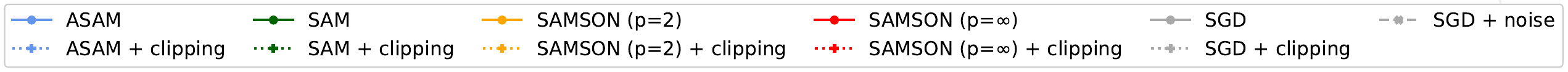}
     \end{subfigure}
     \caption{Performance of the different methods under a range of random conductance variations. We plot the mean and standard deviation over 10 and 3 inference runs for CIFAR-10/100 and ImageNet, respectively.}
     \label{fig:conductance_variation_AdaBS}
\end{figure*}

\section{Model robustness on a generic noise model}\label{sec:genericnoise}

We will now focus on analyzing how sharpness-aware training promotes DNN robustness compared to standard SGD training. In particular, we will focus on improving robustness in the context of noisy hardware accelerators that exploit the energy-reliability trade-off to improve energy efficiency at the cost of noisy weights. As our use-case, we consider memristor-based DNN implementations, which present a promising direction in energy-efficient DNN inference accelerators~\cite{joshi2020accurate,kern2022memse}. In such a setting, the weights of all fully-connected or convolutional layers of a pre-trained DNN are linearly mapped to the range of possible conductance values from 0 to $G_{\text{max}}$. More concretely, the ideal conductance values $G^l_{T,ij}$ for the weights $W^l_{ij}$ of layer $l$ is
\begin{equation}
    G^l_{T,ij} = W^l_{ij} \times \frac{G_{\text{max}}}{W^l_{\text{max}}}
\end{equation}
where $W^l_{\text{max}}$ is layer $l$'s maximum absolute weight. However, as pointed out previously, $G^l_{T,ij}$ is not achievable in practice since conductance errors $\delta_{ij}$ are originated from programming and read noise~\cite{programming_read_noise} as well as conductance drift over time~\cite{conductance_drift}. Hence, in the general case, the non-ideal conductance values $G^l_{ij}$ may be defined as
\begin{equation}
    G^l_{ij} = G^l_{T,ij} \times \delta_{ij}
\label{eq:generic_noise_model}
\end{equation}
with $\delta_{ij} \sim \mathcal{N}(1,\sigma_c^2)$. Following Joshi \etal \cite{joshi2020accurate}, $\sigma_c$ represents the conductance variation of $G^l_{ij}$ relative to $G^l_{T,ij}$. The generic noise model presented in \cref{eq:generic_noise_model} may be used to accurately estimate inference accuracy in noise models derived from measurements of existing noisy hardware implementations. 

We tested robustness in a variety of networks -- VGG-13 trained on CIFAR-10, MobileNetV2 trained on CIFAR-100, and ResNet-18 finetuned on ImageNet -- following the same training procedure describe above.
We tried a range of neighborhood sizes for the various methods since different a $\rho$ provides a distinct trade-off between performance and robustness. Additional details are provided in the appendix. Overall, we found that $\rho = 0.5$ or $\rho = 1.0$ tend to provide the best trade-offs for both SAMSON and ASAM and $\rho = 0.05$ or $\rho = 0.1$ for SAM.
To promote a cleaner visualization, we only report the best $\rho$ for each method. Lastly, we note that $\sigma_c = 0.0$ in our experiments refers to the special case where no noise is applied to the DNN weights.


\subsection{Baseline robustness methods}

\begin{figure*}[t]
     \centering
     \begin{subfigure}[b]{0.33\textwidth}
         \centering
         \includegraphics[width=\textwidth]{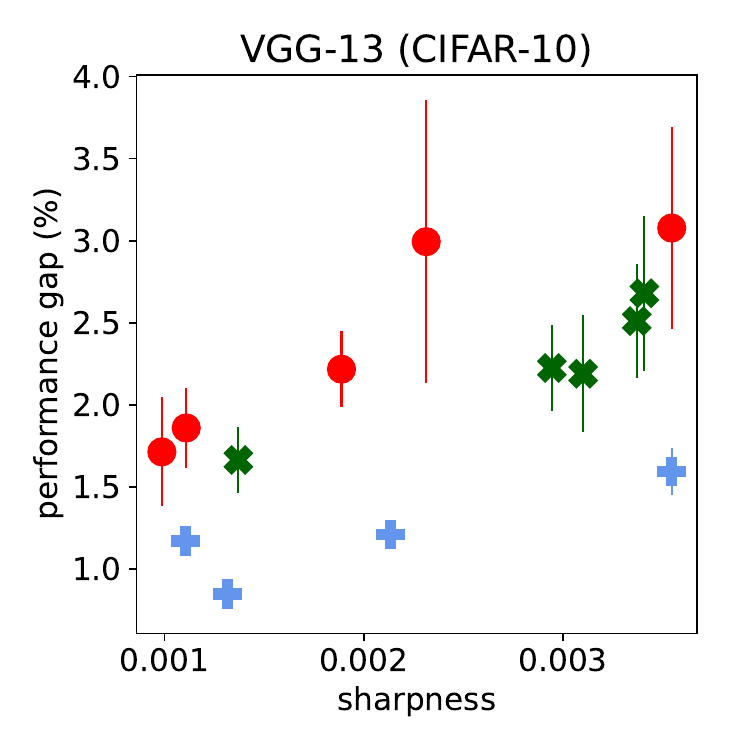}
     \end{subfigure}
     \hfill
     \begin{subfigure}[b]{0.33\textwidth}
         \centering
         \includegraphics[width=\textwidth]{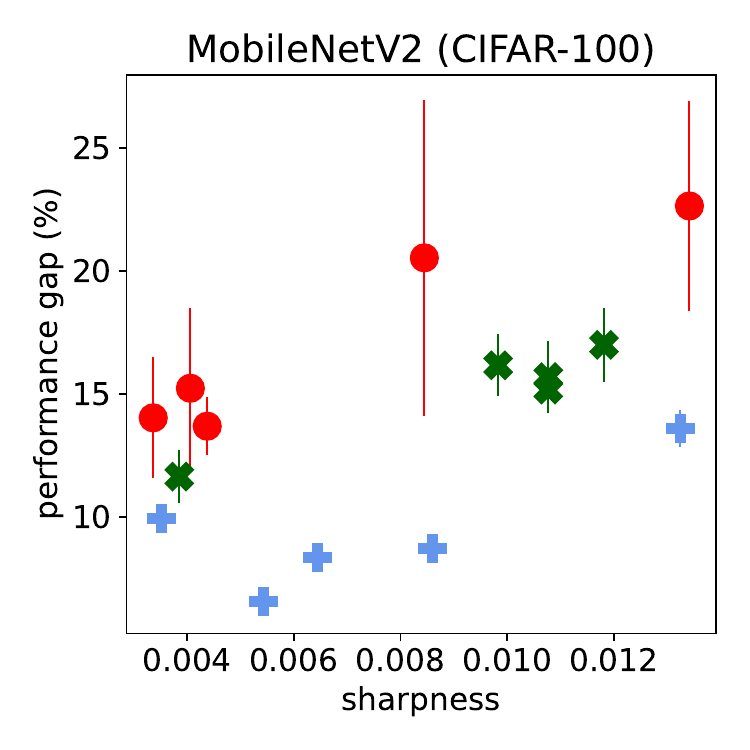}
     \end{subfigure}
     \hfill
     \begin{subfigure}[b]{0.33\textwidth}
         \centering
         \includegraphics[width=\textwidth]{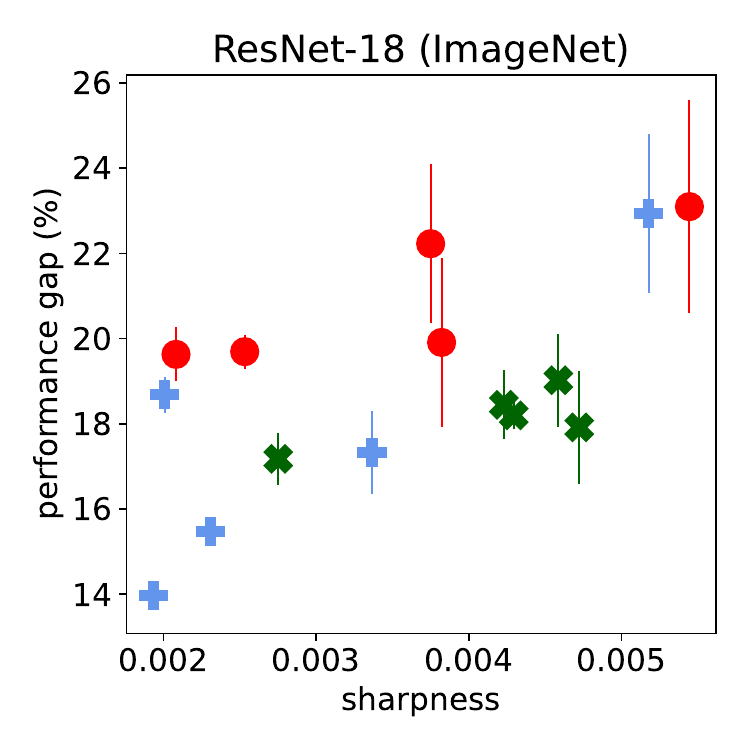}
     \end{subfigure}
     \begin{subfigure}[b]{0.99\textwidth}
        \centering
        \includegraphics[width=0.4\textwidth]{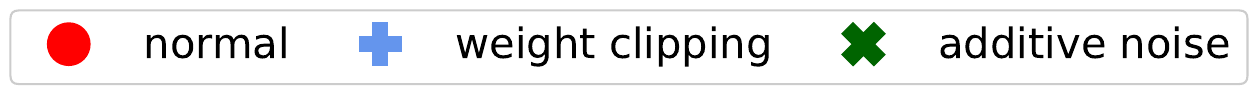}
     \end{subfigure}
     \caption{Correlation between SAMSON$_2$'s $m$-sharpness (\cref{eq:m_sharpness_samson}, $\rho=0.5$, $p=2$) and robustness, \ie the performance gap between the noise realizations at $\sigma_c=0.0$ and at $\sigma_c=0.4$. We plot the mean and standard deviation over 10 and 3 inference runs for CIFAR-10/100 and ImageNet, respectively}
     \label{fig:sharpness_performance_AdaBS_samson_2}
\end{figure*}

\begin{figure*}[t]
     \centering
     \begin{subfigure}[b]{0.33\textwidth}
         \centering
         \includegraphics[width=\textwidth]{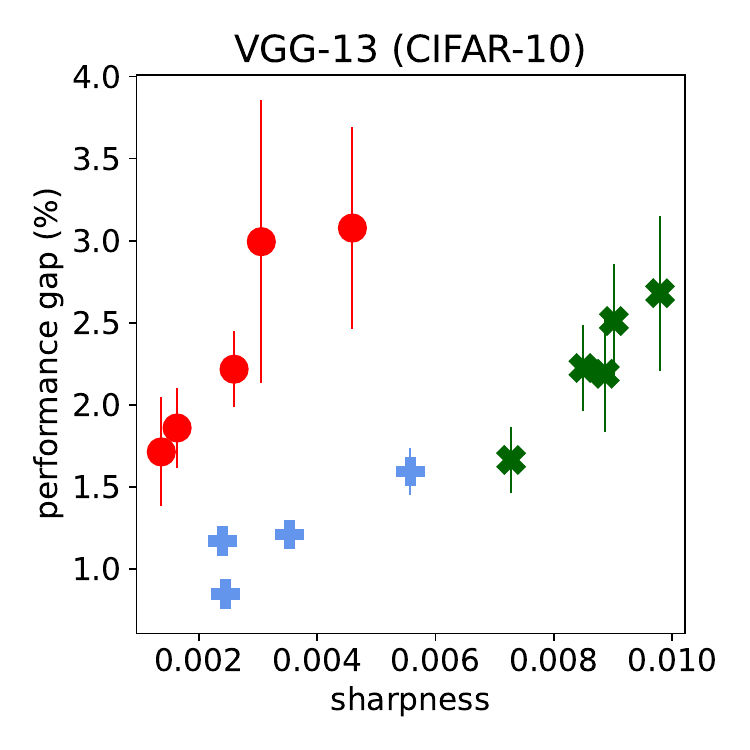}
     \end{subfigure}
     \hfill
     \begin{subfigure}[b]{0.33\textwidth}
         \centering
         \includegraphics[width=\textwidth]{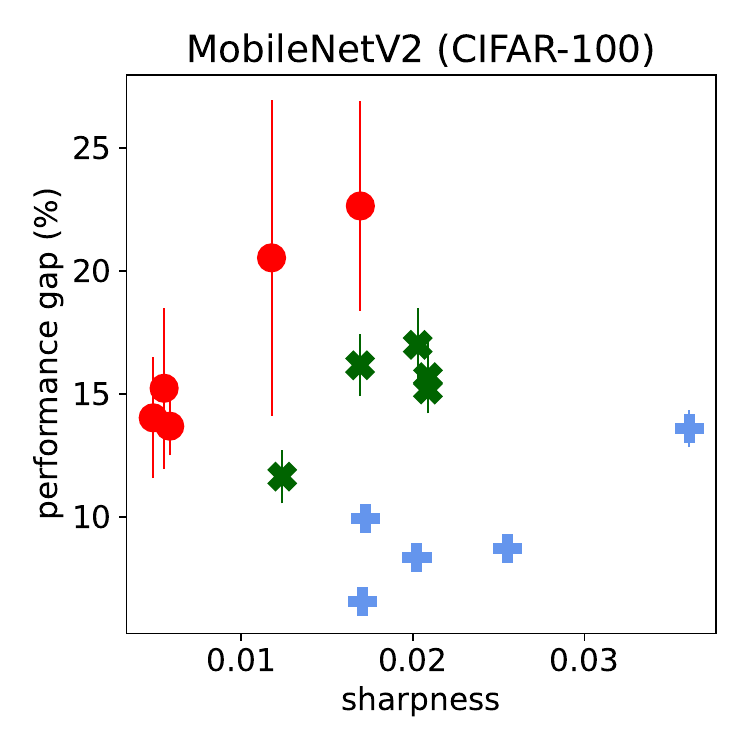}
     \end{subfigure}
     \hfill
     \begin{subfigure}[b]{0.33\textwidth}
         \centering
         \includegraphics[width=\textwidth]{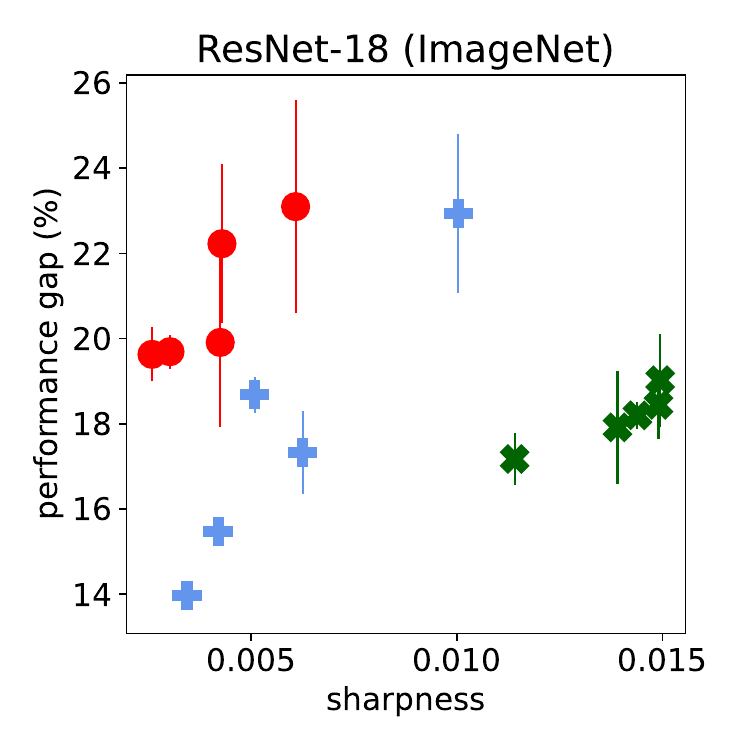}
     \end{subfigure}
     \begin{subfigure}[b]{0.99\textwidth}
        \centering
        \includegraphics[width=0.4\textwidth]{Figures/sharpness_legend.pdf}
     \end{subfigure}
     \caption{Correlation between SAMSON$_\infty$'s $m$-sharpness (\cref{eq:m_sharpness_samson}, $\rho=0.5$, $p=\infty$) and robustness, \ie the performance gap between the noise realizations at $\sigma_c=0.0$ and at $\sigma_c=0.4$. We plot the mean and standard deviation over 10 and 3 inference runs for CIFAR-10/100 and ImageNet, respectively.}
     \label{fig:sharpness_performance_AdaBS_samson_inf}
\end{figure*}

On top of a simple baseline trained with vanilla SGD, we experimented with two methods: the additive noise approach proposed by Joshi \etal \cite{joshi2020accurate} and aggressive weight clipping \cite{stutz2021bit}. More specifically, the first method applies additive Gaussian noise to DNN weights, whereas the second method clips the DNN weights into a small range of possible values. The models are trained from scratch and use the training settings previously described.

The additive random noise proposed by Joshi \etal \cite{joshi2020accurate} is sampled from a Gaussian distribution $\mathcal{N}(0,\sigma_n^2)$, where
\begin{equation}
    \sigma_n = \frac{W^l_\text{max} \times \sigma_G}{G_\text{max}} 
\end{equation}
and $\sigma_G$ represents the standard deviation of hardware non-idealities observed in practice. Both $\sigma_G$ and $G_\text{max}$ are device characteristics that are set to $0.94$ and $25$, respectively, following the empirical measurements on 1 million of phase-change memory devices \cite{joshi2020accurate}. Since the amount of added noise is proportional to the maximum absolute weight value of a given layer, we perform weight clipping after each weight update; we used the range $\big[ -\alpha \times \sigma_{W^l}, \alpha \times \sigma_{W^l} \big]$, where $\sigma_{W^l}$ is the standard deviation of the weights of layer $l$ and $\alpha$ is a predefined hyper-parameter defaulted to $2.0$. We tried a different range of $\alpha \in \{1.5, 2.0, 2.5\}$, but the best performance for all CIFAR-10/100 models was achieved with the default $\alpha$ value of $2.0$. For finetuning on ImageNet, we used $\alpha$ = 2.5, as the original authors suggested \cite{joshi2020accurate}.

For aggressive weight clipping, we tried the values for the clipping range $c$, as performed by the original authors \cite{stutz2021bit}: $\{\pm0.05, \pm0.10, \pm0.15, \pm0.20\}$. A lower weight range induced by a smaller $c$ leads to highly robust networks. However, they may lack generalization performance in the noiseless to low-noise regimes due to outlier distortion. Hence, manipulating $c$ provides a trade-off between performance and robustness. In our experiments, we observed that 0.2 (and in some cases 0.15) achieved the best trade-off and was used on most of the reported networks. Please see the appendix for additional details.

To reduce the impact of hardware non-idealities in the DNN performance, Joshi \etal \cite{joshi2020accurate} also proposed adaptive batch normalization statistics (AdaBS), which updates the batch normalization statistics using a calibration set. More specifically, the running mean and running variance of all batch normalization layers are updated using the statistics computed during inference on a calibration set using noisy weights. AdaBS was shown to significantly retain performance at high-noise settings, compared to using simpler techniques such as global drift compensation~\cite{le2018compressed}. We used the originally suggested hyper-parameters and applied AdaBS in all the networks used in our experiments.

\subsection{Robustness to different conductance variation}



The robustness of the models trained with SAMSON, ASAM, SAM, and SGD in combination with aggressive weight clipping at different conductance variation levels is shown in \cref{fig:conductance_variation_AdaBS}. We also include training with SGD and additive Gaussian noise as an additional baseline. For visualization clarity, we exclude training with additive noise on top of the sharpness-aware training variants but refer to the appendix for these results. 

We observe that SAMSON variants primarily compose the Pareto frontier across all models and datasets. Ultimately, this means that training a DNN with SAMSON with and without aggressive weight clipping provides the best performance and robustness trade-off across all noisy regimes. This is also observed in the noiseless regime ($\sigma_c = 0.0$), where we see that there is always at least one SAMSON variant that achieves the best test accuracy, as discussed in section~\ref{sec:generalization_performance}. The difference in model robustness between the various methods is more subtle on ImageNet, likely due to all methods starting with the same pre-trained model and being only finetuned for 10 epochs. Nevertheless, we see that SAMSON is the only method able to provide significant improvements in terms of robustness in highly noisy regimes, \textit{e.g.} $\sigma_c = 0.4$. 

Overall, we observe that sharpness-aware training variants (SAMSON, ASAM, and SAM) clearly outperform SGD, with SAMSON promoting the highest robustness, generally followed by ASAM and then SAM. This is seen in terms of not only robustness at different noise levels but also in the best performances achieved in the noiseless regime. Moreover, the improvement in robustness is especially amplified when combining sharpness-aware methods with aggressive weight clipping, representing a simple yet effective alternative to training with noise. We note that, as expected, the performance on the clean network drops when applying both weight clipping or additive noise, as observed in the zoomed-in patches. This ultimately mitigates the robustness benefits while using these methods in lower noisy settings but proves to be remarkably beneficial in highly noisy regimes.

\subsection{Correlation between sharpness and robustness}

For measuring sharpness, we use the $m$-sharpness metric proposed by Foret \etal \cite{foret2021sharpnessaware}, which stems from the original SAM formulation (\cref{eq:sam}), and further extend it to SAMSON's objective (\cref{eq:samson}). Considering a training set ($S_{\text{train}}$) composed of $n$ minibatches $S$ of size $m$, we compute the difference of the loss $l_s$ of a given sample $s$ with and without a worst-case perturbation $\epsilon$ on $w$. SAMSON's $m$-sharpness is calculated as
\begin{equation}
    \dfrac{1}{n} \sum_{S \in S_{\text{train}}} \max_{\|\epsilon \|\pmb{w}\|^{-1}_p / |w| \|_2 \leq \rho} \dfrac{1}{m} \sum_{s \in S} l_s(w + \epsilon) - l_s(w).
    \label{eq:m_sharpness_samson}
\end{equation}
In our experiments, we used $m = 400$ and $m = 128$ for measuring the sharpness of models finetuned on ImageNet and trained on CIFAR-10/100, respectively.

We treat robustness as the performance gap measured by the difference in test accuracy between the noiseless models, \textit{i.e.} with no conductance variation applied to the weights ($\sigma_c = 0.0$), and the noisy model configurations with the highest tested conductance variation ($\sigma_c=0.4$). We present the relation between sharpness and robustness of all the tested models using SAMSON's $m$-sharpness with $p=2$ and $p=\infty$ in figs. \ref{fig:sharpness_performance_AdaBS_samson_2} and \ref{fig:sharpness_performance_AdaBS_samson_inf}, respectively.

We observe a strong correlation within each training configuration, \textit{i.e.} training each method with and without additive noise or aggressive weight clipping, across all architectures and datasets. Such results showcase the ability of SAMSON's $m$-sharpness in acting as a generic robustness metric. Importantly, this suggests that training with SAMSON's objective, especially when combined with existing robustness methods such as aggressive weight clipping, is an effective way of promoting more robust DNNs at inference time.

We provide visualizations of other sharpness metrics in the appendix, particularly $m$-sharpness as calculated using SAM's and ASAM's objectives and the metric proposed by Keskar \etal~\cite{keskar2016large}, which is computed based on the largest loss value around a cuboid neighborhood shape. We note that ASAM's $m$-sharpness particularly shows a strong correlation pattern, followed by SAM's $m$-sharpness. This further suggests that sharpness-aware training is a good proxy for increasing robustness in noisy hardware settings, matching our previous findings regarding the additional robustness achieved by SAMSON, ASAM, and SAM variants when compared to vanilla SGD.

\section{Model robustness on noise simulations from real hardware}

Since the generic noise model used in section~\ref{sec:genericnoise} may not exactly match existing hardware implementations, we also performed experiments using an inference simulator on real hardware provided by IBM's analog hardware acceleration kit~\cite{rasch2021flexible}. This simulator uses the empirical measurements from 1 million phase-change memory devices \cite{nandakumar2019phase} to accurately simulate how hardware noise affects the DNN weights \cite{joshi2020accurate}. Specifically, by taking into account the programming and read noise, we can simulate the DNN weights of a given model after deployment. In particular, we can measure how a model would perform after a specific amount of time has passed since its weights were transferred to the target hardware. We refer to the library's documentation\footnote[2]{\url{https://aihwkit.readthedocs.io/en/latest/pcm_inference.html}} for additional details.




\begin{figure}[t]
     \centering
     \includegraphics[width=0.48\textwidth]{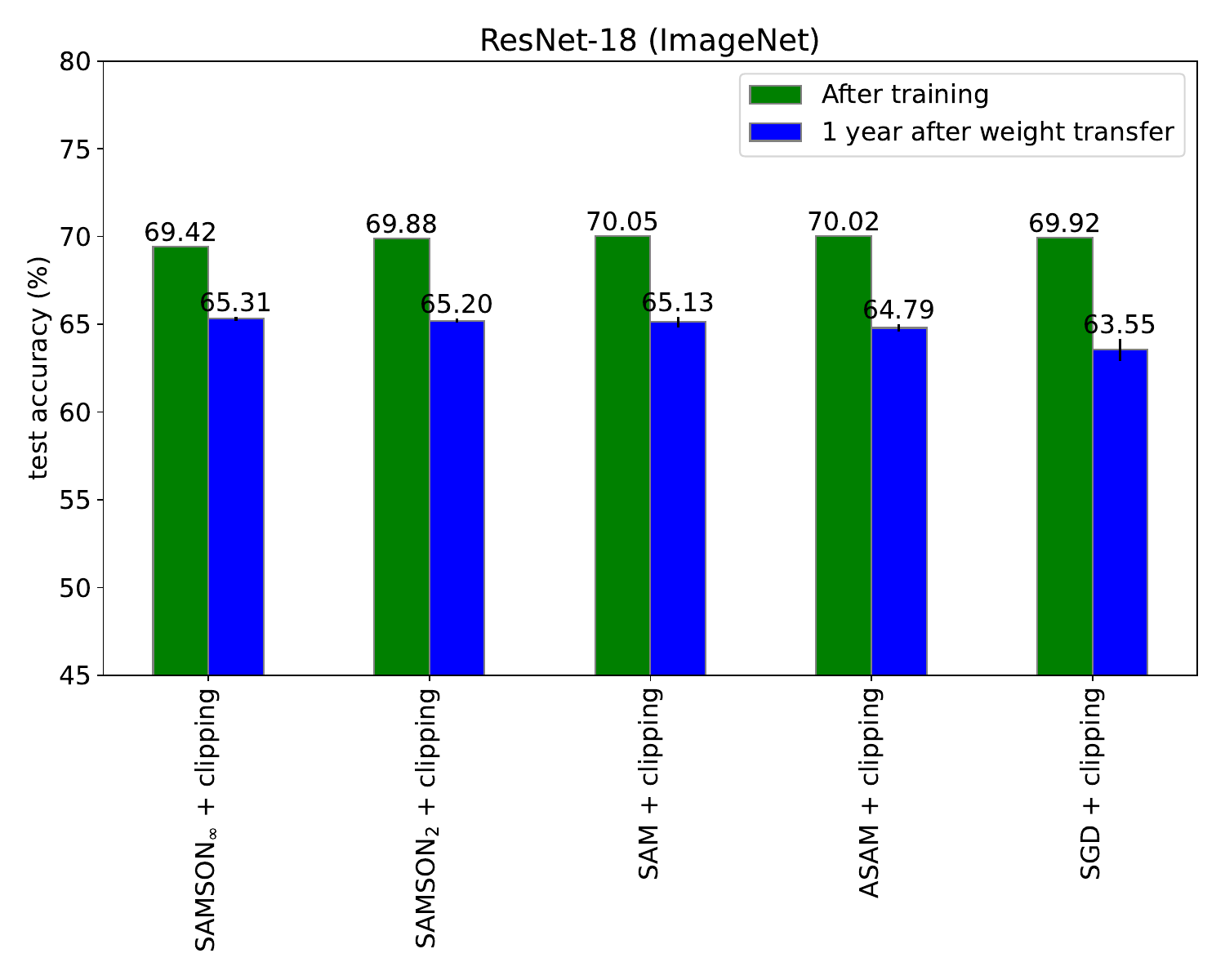}
     \caption{Performance of the different methods with aggressive weight clipping on ResNet-18 finetuned on ImageNet 1 year after weight transfer to the target hardware. We plot the mean and standard deviation over 10 inference runs.}
     \label{fig:weight_transfer_imagenet_finetuned}
\end{figure}

\subsection{Performance after weight transfer}



The main goal of this experiment is to provide a practical example of how the performance on the generic noise model relates to existing hardware implementations. We report the performance of the different methods combined with aggressive weight clipping measured 1 year after deployment on the target hardware. The performance and robustness of the different methods using ResNet-18 models finetuned on ImageNet are presented in \cref{fig:weight_transfer_imagenet_finetuned}. 

We observe that even though all sharpness-aware training methods outperform SGD in terms of robustness, the SAMSON variants retain the most performance. This is particularly important in scenarios where often reprogramming the DNN weights on the memristor device is not practical or feasible. For example, the originally programmed device might no longer be reachable for reprogramming after it is shipped to a client or a remote destination. In such cases, retaining as much performance as possible over time is an important requisite of the deployed DNN.

\section{Conclusion}

In this work, we propose a new adaptive sharpness-aware training method that conditions the individual worst-case perturbation of a given weight based on not only its absolute value but also on the weight range distribution of a particular layer. Our results on different architectures, datasets, and training regimes showcase the benefits of training with our approach, SAMSON, to increase DNN performance in noiseless settings. Moreover, we demonstrate that SAMSON in combination with aggressive weight clipping is an effective way to improve DNN robustness to weight noise. Furthermore, we show a high correlation between SAMSON's objective and model robustness. Our method and consequent findings provide crucial steps to achieving highly energy-efficient DNN systems at inference time by enabling the exploitation of the energy-reliability trade-off of existing DNN accelerators.



{\small
\bibliographystyle{ieee_fullname}
\bibliography{egbib}
}

\clearpage
\appendix

\begin{figure*}[ht]
     \centering
     \begin{subfigure}[b]{0.33\textwidth}
         \centering
         \includegraphics[width=\textwidth]{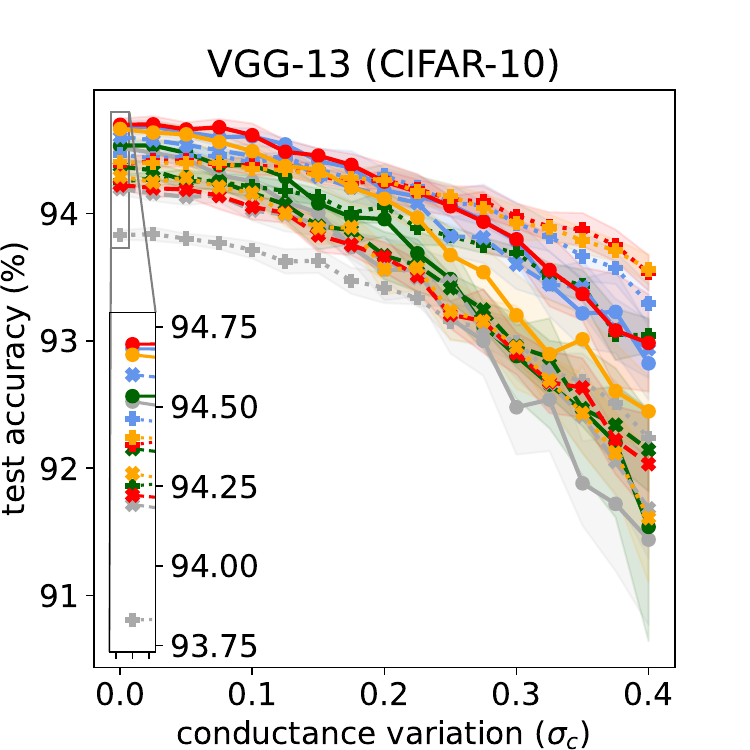}
     \end{subfigure}
     \hfill
     \begin{subfigure}[b]{0.33\textwidth}
         \centering
         \includegraphics[width=\textwidth]{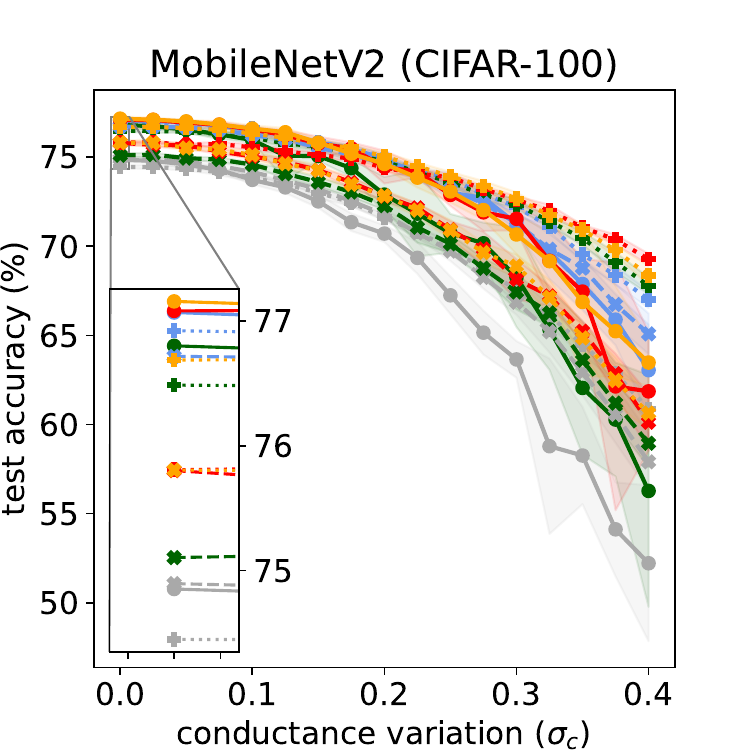}
     \end{subfigure}
     \hfill
     \begin{subfigure}[b]{0.33\textwidth}
         \centering
         \includegraphics[width=\textwidth]{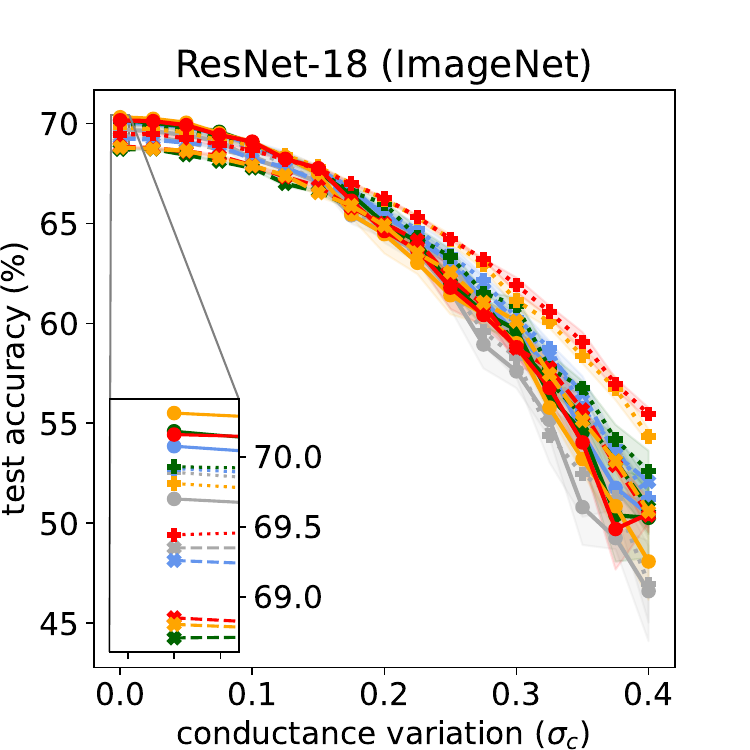}
     \end{subfigure}
     \hfill
     \begin{subfigure}[b]{0.99\textwidth}
        \centering
        \includegraphics[width=\textwidth]{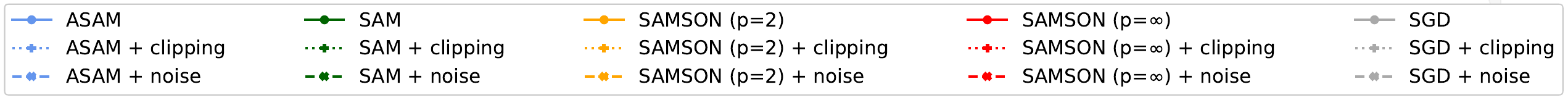}
     \end{subfigure}
     \caption{Performance of the different methods under a range of random conductance variations and combined with either weight clipping or additive noise. We plot the mean and standard deviation over 10 and 3 inference runs for CIFAR-10/100 and ImageNet, respectively.}
     \label{fig:conductance_variation_AdaBS_with_additive_noise}
\end{figure*}

\begin{figure*}[t]
     \centering
     \begin{subfigure}[b]{0.33\textwidth}
         \centering
         \includegraphics[width=\textwidth]{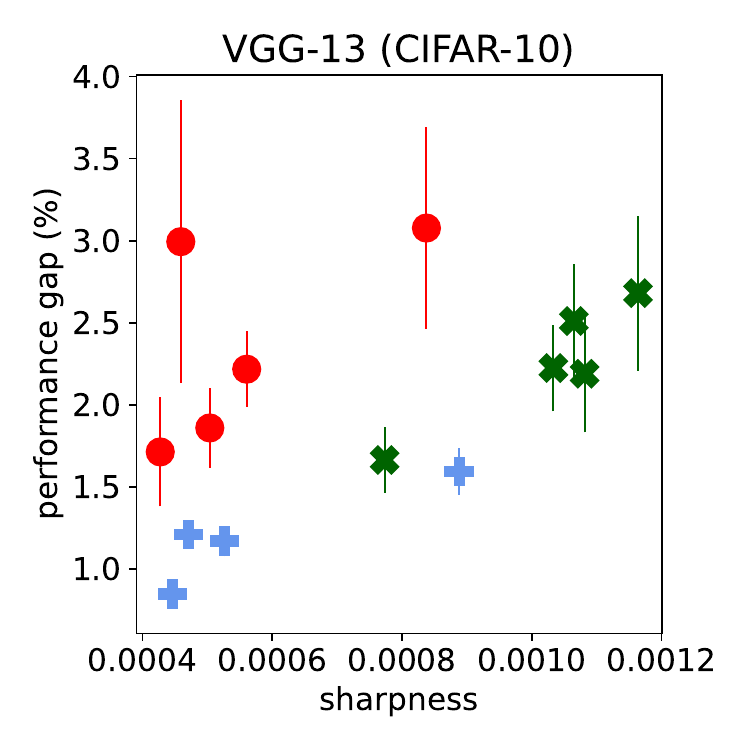}
     \end{subfigure}
     \hfill
     \begin{subfigure}[b]{0.33\textwidth}
         \centering
         \includegraphics[width=\textwidth]{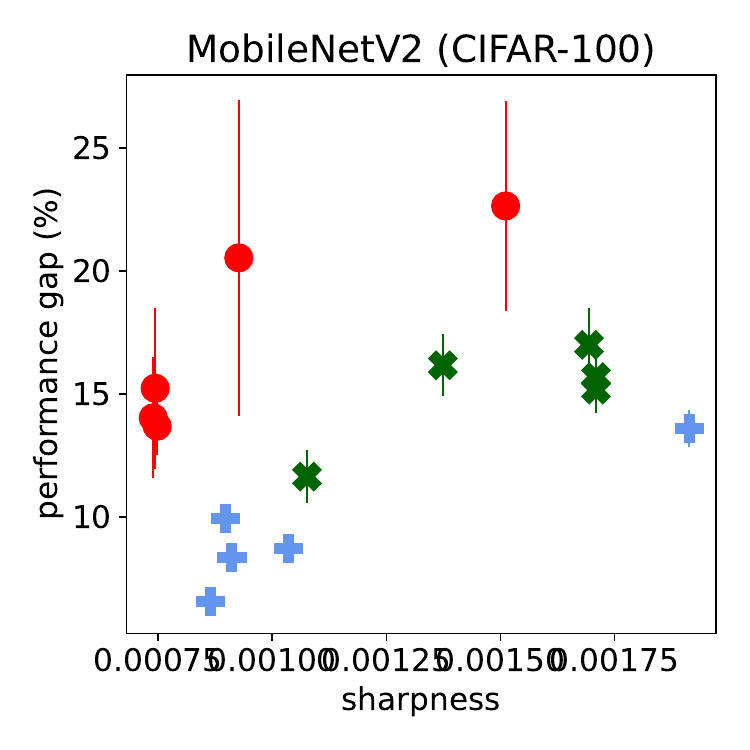}
     \end{subfigure}
     \hfill
     \begin{subfigure}[b]{0.33\textwidth}
         \centering
         \includegraphics[width=\textwidth]{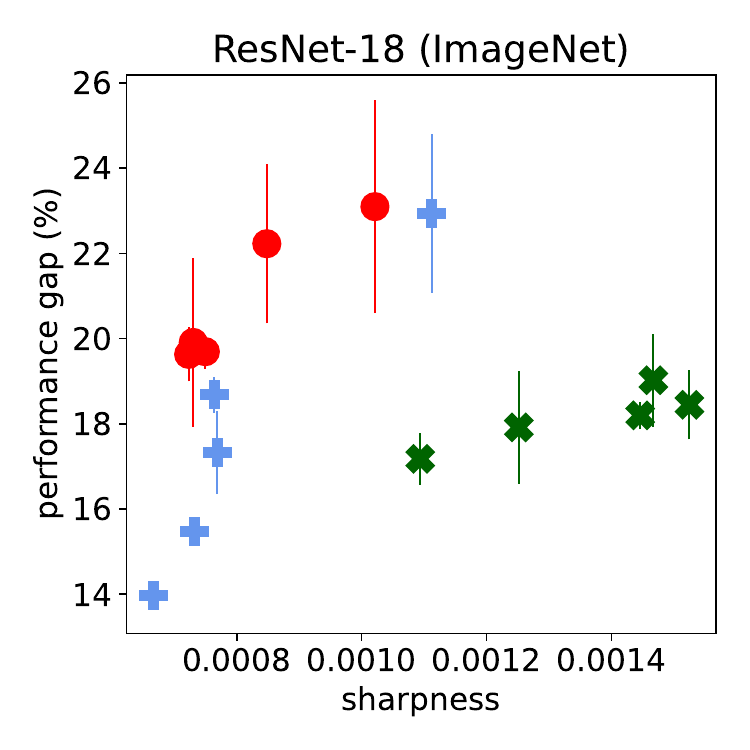}
     \end{subfigure}
     \begin{subfigure}[b]{0.99\textwidth}
        \centering
        \includegraphics[width=0.4\textwidth]{Figures/sharpness_legend.pdf}
     \end{subfigure}
     \caption{Correlation between SAM's $m$-sharpness (\cref{eq:m_sharpness_sam}, $\rho=0.05$) and robustness, \ie the performance gap between the noise realizations at $\sigma_c=0.0$ and at $\sigma_c=0.4$. We plot the mean and standard deviation over 10 inference runs.}
     \label{fig:sharpness_performance_AdaBS_sam}
\end{figure*}

\begin{figure*}[t]
     \centering
     \begin{subfigure}[b]{0.33\textwidth}
         \centering
         \includegraphics[width=\textwidth]{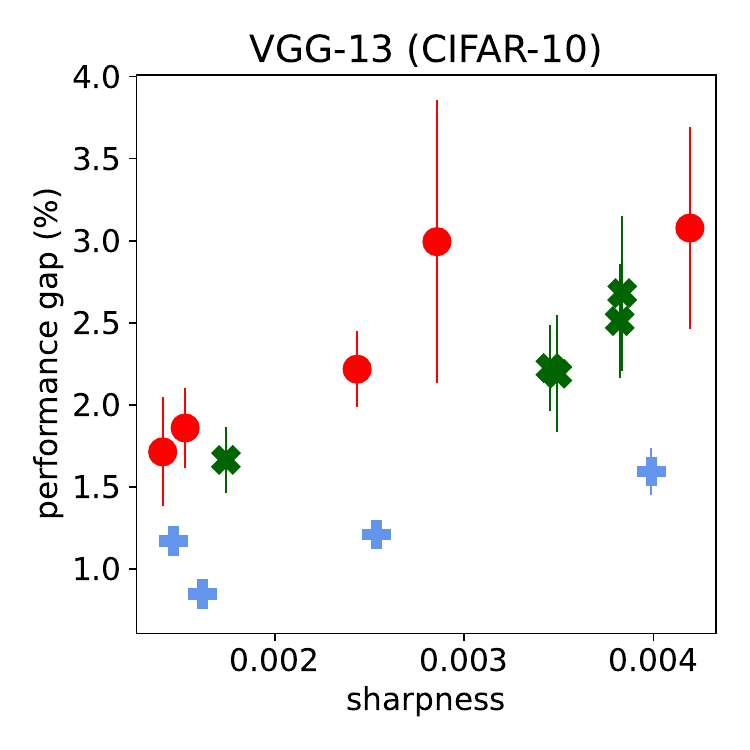}
     \end{subfigure}
     \hfill
     \begin{subfigure}[b]{0.33\textwidth}
         \centering
         \includegraphics[width=\textwidth]{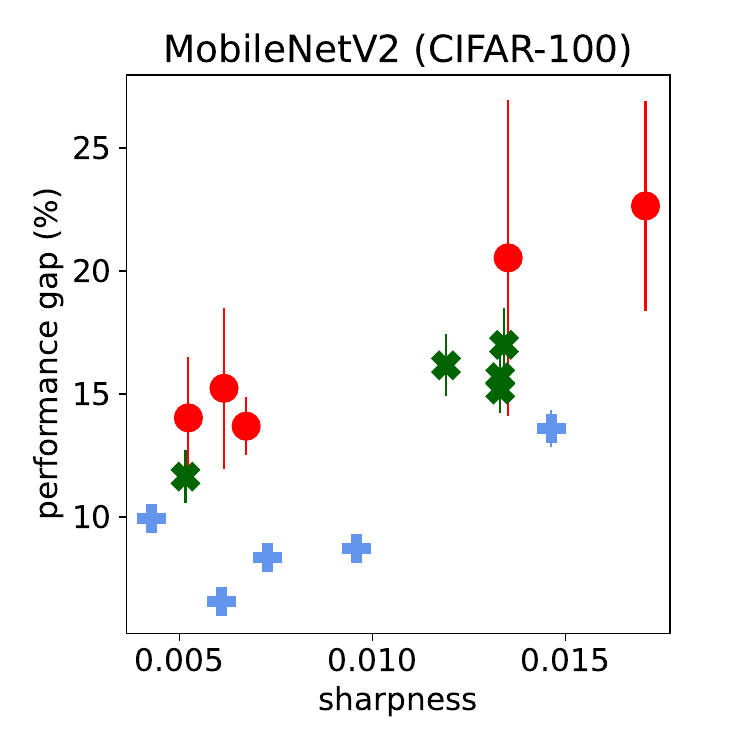}
     \end{subfigure}
     \hfill
     \begin{subfigure}[b]{0.33\textwidth}
         \centering
         \includegraphics[width=\textwidth]{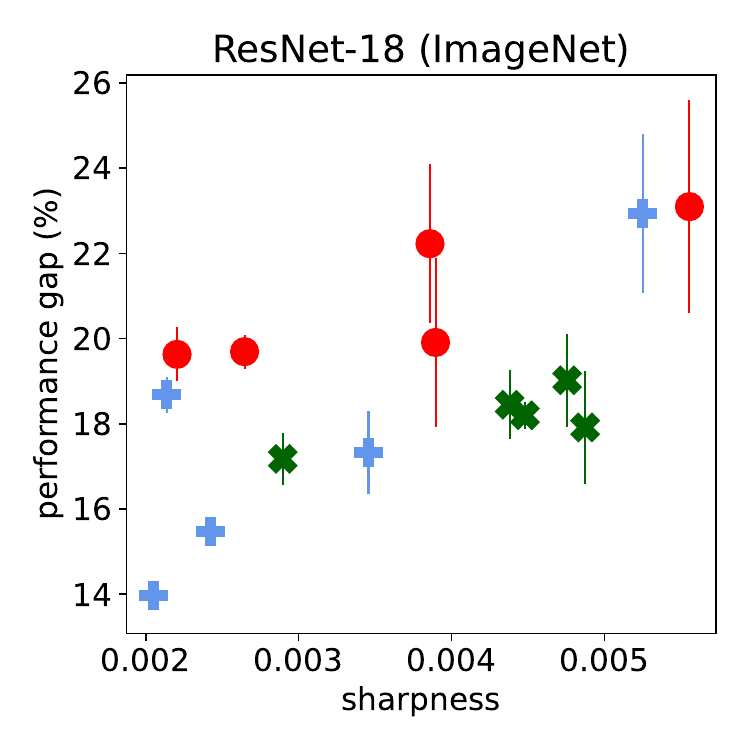}
     \end{subfigure}
     \begin{subfigure}[b]{0.99\textwidth}
        \centering
        \includegraphics[width=0.4\textwidth]{Figures/sharpness_legend.pdf}
     \end{subfigure}
     \caption{Correlation between ASAM's $m$-sharpness (\cref{eq:m_sharpness_asam}, $\rho=0.5$) and robustness, \ie the performance gap between the noise realizations at $\sigma_c=0.0$ and at $\sigma_c=0.4$. We plot the mean and standard deviation over 10 inference runs.}
     \label{fig:sharpness_performance_AdaBS_asam}
\end{figure*}

\begin{figure*}[t]
     \centering
     \begin{subfigure}[b]{0.33\textwidth}
         \centering
         \includegraphics[width=\textwidth]{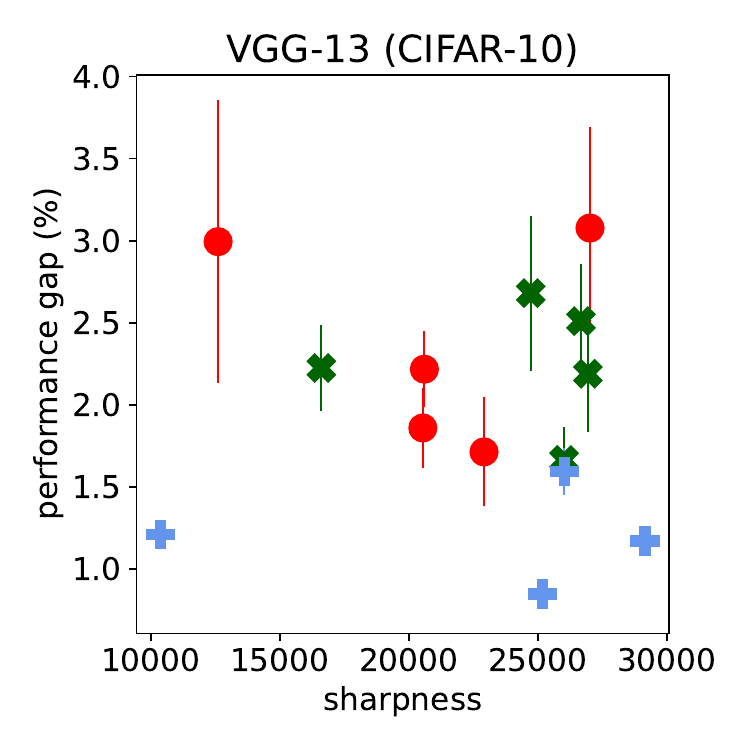}
     \end{subfigure}
     \hfill
     \begin{subfigure}[b]{0.33\textwidth}
         \centering
         \includegraphics[width=\textwidth]{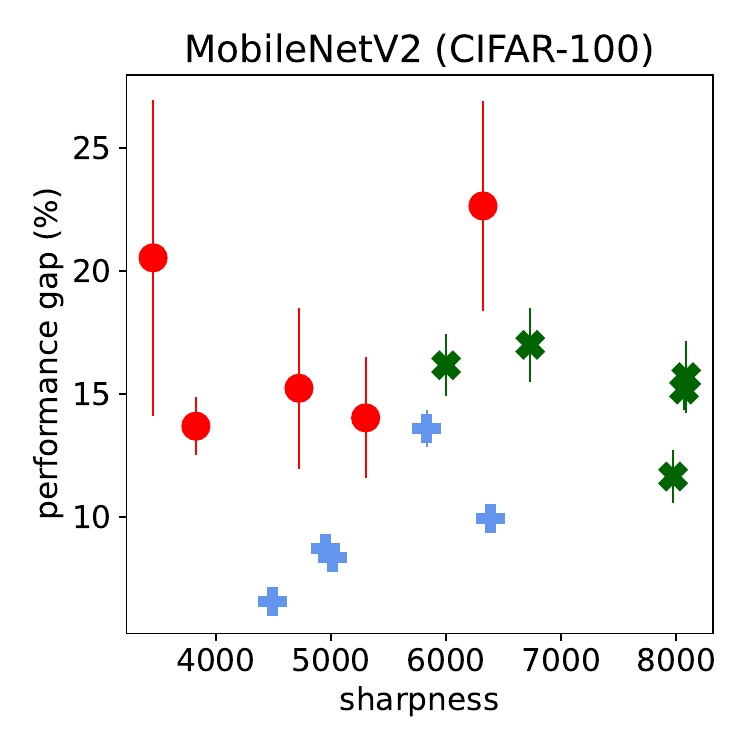}
     \end{subfigure}
     \hfill
     \begin{subfigure}[b]{0.33\textwidth}
         \centering
         \includegraphics[width=\textwidth]{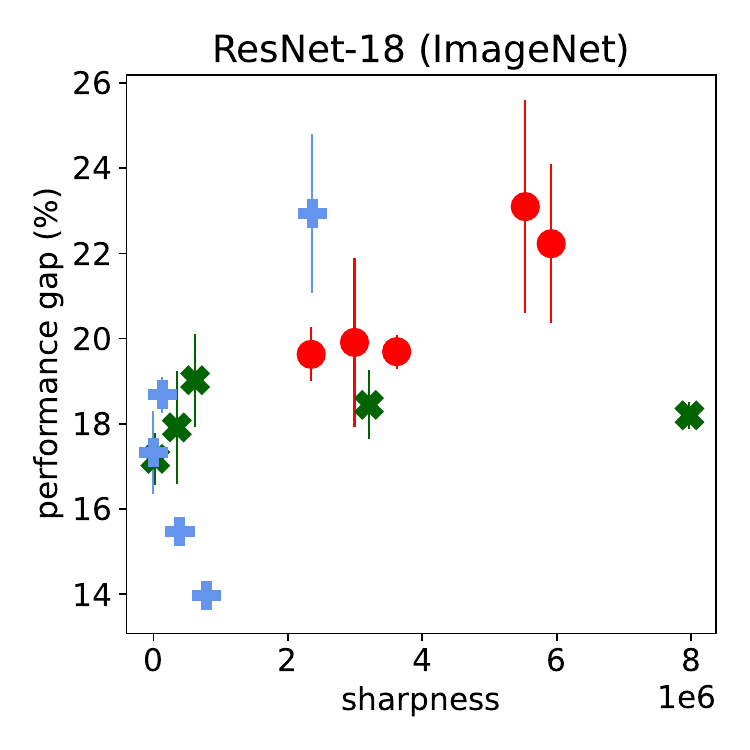}
     \end{subfigure}
     \begin{subfigure}[b]{0.99\textwidth}
        \centering
        \includegraphics[width=0.4\textwidth]{Figures/sharpness_legend.pdf}
     \end{subfigure}
     \caption{Correlation between Keskar \etal \cite{keskar2016large}'s sharpness and robustness, \ie the performance gap between the noise realizations at $\sigma_c=0.0$ and at $\sigma_c=0.4$. We plot the mean and standard deviation over 10 inference runs.}
     \label{fig:sharpness_performance_AdaBS_1e3}
\end{figure*}

\section{Additional robustness experiments}

We also present the robustness results when combining the sharpness-aware training variants (SAM, ASAM, and SAMSON) with additive Gaussian noise in \cref{fig:conductance_variation_AdaBS_with_additive_noise}. Even though we observe an increase in robustness in certain configurations, training with aggressive weight clipping tends to provide the overall best trade-off between performance and robustness compared to training with additive noise.

\section{Hyper-parameter tuning}
\label{app:hp_tuning}

The considered ranges for the different hyper-parameters are presented in \cref{tab:hyperparam_choices}. The configurations with the best performance and robustness trade-off for the models trained CIFAR-10, CIFAR-100, and ImageNet are presented in tables \ref{tab:hyperparam_cifar10}, \ref{tab:hyperparam_cifar100}, and \ref{tab:hyperparam_imagenet}, respectively. These configurations were the ones used to report the results in the main paper.

\begin{table}[ht]
\begin{center}
\begin{small}
\begin{sc}
\begin{tabular}{cc}
\toprule
Hyper-parameter & Choices \\
\midrule
SAM's $\rho$ & \{0.05, 0.1, 0.2, 0.5\} \\
ASAM's $\rho$ & \{0.5, 1.0, 1.5, 2.0\} \\
SAMSON's $p$ & \{2, $\infty$\} \\
SAMSON's $\rho$ & \{0.1, 0.2, 0.5, 1.0\} \\
$c$ & \{$\pm$0.05, $\pm$0.10, $\pm$0.15, $\pm$0.20\} \\
$\alpha$ & \{1.5, 2.0, 2.5\}\\
\bottomrule
\end{tabular}
\end{sc}
\end{small}
\end{center}
\caption{Hyper-parameter choices for the different methods.}
\label{tab:hyperparam_choices}
\end{table}

\begin{table}[ht]
\begin{center}
\begin{small}
\begin{sc}
\begin{tabular}{lr}
\toprule
Method & Best configuration \\
\midrule
SGD + noise & $\alpha = 2.0$\\
SGD + clipping & $c = \pm 0.15$ \\
SAM & $\rho = 0.1$\\
SAM + noise & $\rho = 0.1, \alpha = 2.0$\\
SAM + clipping & $\rho = 0.1, c = \pm 0.2$\\
ASAM & $\rho = 0.5$\\
ASAM + noise & $\rho = 0.5, \alpha = 2.0$\\
ASAM + clipping & $\rho = 0.5, c = \pm 0.2$\\
SAMSON$_2$ & $\rho = 0.2, p = 2$\\
SAMSON$_2$ + clipping & $\rho = 0.5, p = 2, c = \pm 0.2$\\
SAMSON$_2$ + noise & $\rho = 0.1, p = 2, \alpha = 2.5$\\
SAMSON$_\infty$ & $\rho = 1.0, p = \infty$\\
SAMSON$_\infty$ + clipping & $\rho = 0.5, p = \infty, c = \pm 0.2$\\
SAMSON$_\infty$ + noise & $\rho = 0.1, p = \infty, \alpha = 2.5$\\
\bottomrule
\end{tabular}
\end{sc}
\end{small}
\end{center}
\caption{Best hyper-parameter configurations for VGG-13 trained on CIFAR-10.}
\label{tab:hyperparam_cifar10}
\end{table}

\begin{table}[ht]
\begin{center}
\begin{small}
\begin{sc}
\begin{tabular}{lr}
\toprule
Method & Best configuration \\
\midrule
SGD + noise & $\alpha = 2.0$\\
SGD + clipping & $c = \pm 0.2$ \\
SAM & $\rho = 0.2$\\
SAM + noise & $\rho = 0.2, \alpha = 2.0$\\
SAM + clipping & $\rho = 0.2, c = \pm 0.2$\\
ASAM & $\rho = 1.0$\\
ASAM + noise & $\rho = 1.0, \alpha = 2.0$\\
ASAM + clipping & $\rho = 1.0, c = \pm 0.2$\\
SAMSON$_2$ & $\rho = 1.0, p = 2$\\
SAMSON$_2$ + clipping & $\rho = 0.5, p = 2, c = \pm 0.2$\\
SAMSON$_2$ + noise & $\rho = 0.2, p = 2, \alpha = 2.5$\\
SAMSON$_\infty$ & $\rho = 1.0, p = \infty$\\
SAMSON$_\infty$ + clipping & $\rho = 1.0, p = \infty, c = \pm 0.2$\\
SAMSON$_\infty$ + noise & $\rho = 0.2, p = \infty, \alpha = 2.5$\\
\bottomrule
\end{tabular}
\end{sc}
\end{small}
\end{center}
\caption{Best hyper-parameter configurations for MobileNetV2 trained on CIFAR-100.}
\label{tab:hyperparam_cifar100}
\end{table}

\begin{table}[ht]
\begin{center}
\begin{small}
\begin{sc}
\begin{tabular}{lr}
\toprule
Method & Best config. \\
\midrule
SGD + noise & $\alpha = 2.5$\\
SGD + clipping & $c = \pm 0.2$ \\
SAM & $\rho = 0.1$\\
SAM + noise & $\rho = 0.05, \alpha = 2.5$\\
SAM + clipping & $\rho = 0.1 , c = \pm 0.2 $\\
ASAM & $\rho = 1.0 $\\
ASAM + noise & $\rho = 0.5, \alpha = 2.5$\\
ASAM + clipping & $\rho = 1.0, c =  \pm 0.2$\\
SAMSON$_2$ & $\rho = 0.2, p = 2$\\
SAMSON$_2$ + clipping & $\rho = 0.5, p = 2, c = \pm 0.2$\\
SAMSON$_2$ + noise & $\rho = 0.1, p = 2, \alpha = 2.5$\\
SAMSON$_\infty$ & $\rho = 0.5, p = \infty$\\
SAMSON$_\infty$ + clipping & $\rho = 1.0, p = \infty, c = \pm 0.2$\\
SAMSON$_\infty$ + noise & $\rho = 0.1, p = \infty, \alpha = 2.5$\\
\bottomrule
\end{tabular}
\end{sc}
\end{small}
\end{center}
\caption{Best hyper-parameter configurations for ResNet-18 finetuned on ImageNet.}
\label{tab:hyperparam_imagenet}
\end{table}

\section{Additional sharpness experiments}
\label{app:add_experiments}

We also provide correlation results with additional sharpness metrics. Particularly, we analyze the $m$-sharpness as formulated per SAM and ASAM's objectives. For SAM, $m$-sharpness is calculated as
\begin{equation}
    \dfrac{1}{n} \sum_{S \in S_{\text{train}}} \max_{\|\epsilon\|_2 \leq \rho} \dfrac{1}{m} \sum_{s \in S} l_s(w + \epsilon) - l_s(w),
    \label{eq:m_sharpness_sam}
 \end{equation}
whereas for ASAM, $m$-sharpness is obtained by
\begin{equation}
    \dfrac{1}{n} \sum_{S \in S_{\text{train}}} \max_{\|\epsilon / |w| \|_2 \leq \rho} \dfrac{1}{m} \sum_{s \in S} l_s(w + \epsilon) - l_s(w).
    \label{eq:m_sharpness_asam}
\end{equation}
To avoid repetition, we refer to the main paper for notations.

Visual correlations between loss sharpness and model robustness using SAM and ASAM's $m$-sharpness are presented in figs. \ref{fig:sharpness_performance_AdaBS_sam} and \ref{fig:sharpness_performance_AdaBS_asam}, respectively. Results using Keskar \etal \cite{keskar2016large}'s sharpness are also shown in \cref{fig:sharpness_performance_AdaBS_1e3}. Overall, we observe that both SAM's and ASAM's $m$-sharpness show better visual correlation than Keskar \etal \cite{keskar2016large}'s notion of sharpness. This suggests that optimizing for low sharpness during training by using existing sharpness-aware training methods is an effective way to promote robustness at inference time, as discussed in the main paper.

\end{document}